\pdfoutput=1
\documentclass[10pt,twocolumn,letterpaper]{article}

\usepackage{btas}
\usepackage{times}
\usepackage{epsfig}
\usepackage{graphicx}
\usepackage{amsmath}
\usepackage{amssymb}
\usepackage{threeparttable}
\usepackage[usenames, dvipsnames]{color}

\usepackage{lipsum}
\newcommand\blfootnote[1]{%
  \begingroup
  \renewcommand\thefootnote{}\footnote{#1}%
  \addtocounter{footnote}{-1}%
  \endgroup
}

\btasfinalcopy 

\ifbtasfinal\fi
\begin{document}

\title{When Face Recognition Meets with Deep Learning: an  Evaluation of Convolutional Neural Networks for Face Recognition}

\author{\small Guosheng Hu$^*$$^\clubsuit$, Yongxin Yang$^*$$^\diamondsuit$, Dong Yi$^\spadesuit$, Josef Kittler$^\clubsuit$, William Christmas$^\clubsuit$, Stan Z. Li$^\spadesuit$, Timothy Hospedales$^\diamondsuit$\\
\small Centre for Vision, Speech and Signal Processing, University of Surrey, UK$^\clubsuit$ \\
\small Electronic Engineering and Computer Science, Queen Mary University of London, UK$^\diamondsuit$ \\
\small Center for Biometrics and Security Research \& National Laboratory of Pattern Recognition, Chinese Academy of Sciences, China$^\spadesuit$\\
{\tt\scriptsize \{g.hu,j.kittler,w.christmas\}@surrey.ac.uk,\{yongxin.yang,t.hospedales\}@qmul.ac.uk, \{szli,dyi\}@cbsr.ia.ac.cn}
}


\maketitle
\thispagestyle{empty}

\begin{abstract}
Deep learning\blfootnote{$^*$ These authors contributed equally to this work}, in particular Convolutional Neural Network (CNN), has achieved promising results in face recognition recently.
However, it remains an open question: why CNNs work well and how to design a `good' architecture.
The existing works tend to focus on reporting CNN architectures that work well  for face recognition rather than investigate the reason. 
In this work, we conduct an extensive evaluation of CNN-based face recognition systems (CNN-FRS) on a common ground to make our work easily reproducible. 
Specifically, we  use public database LFW (Labeled Faces in the Wild) to train CNNs, 
unlike most existing CNNs trained on private databases. 
We propose three CNN architectures which are the first reported architectures trained using LFW data.
This paper quantitatively compares the architectures of CNNs and evaluates the effect of different implementation choices. 
We identify several useful properties of CNN-FRS. 
For instance,  the dimensionality of the learned features can be significantly reduced without adverse effect on face recognition accuracy.
In addition, a traditional metric learning method exploiting CNN-learned features is evaluated.
Experiments show two  crucial factors to good CNN-FRS performance are the  fusion of multiple CNNs and metric learning. 
To make our work reproducible, source code and models will be made publicly available.

\end{abstract}

\section{Introduction}
\label{intro}
The conventional face recognition pipeline consists of four stages: 
face detection, face alignment,  feature extraction (or face representation) and classification. 
Perhaps the single most important stage is feature extraction. 
In constrained environments, the hand-crafted features such as Local Binary Patterns (LBP) ~\cite{LBP} and 
Local Phase Quantisation (LPQ)~\cite{LPQ,LPQ_HO} have achieved respectable face recognition performance. 
However, the performance using these features  degrades dramatically in 
unconstrained environments where face images cover complex and large intra-personal variations 
such as pose, illumination, expression and occlusion. 
It remains an open problem to find an ideal facial feature which is robust for 
face recognition in unconstrained environments (FRUE).
In the last three years,  convolutional neural network (CNN) rebranded as `deep learning' 
has  achieved very impressive results on FRUE. 
Unlike the traditional hand-crafted features, the CNN learning-based features are more robust to complex intra-personal variations.
More notably, the top three face recognition rates reported on the FRUE benchmark database LFW (Labeled Faces in the Wild) ~\cite{LFW}
have been achieved by CNN methods~\cite{FACEPP2015, DEEPID2P, DEEPID2}.
The success of the latest CNNs on FRUE and more general object recognition task~\cite{AlexNet, PRELU, BatchNormalisation} 
stems from the following facts: (1) much larger labeled training  sets are available; 
(2) GPU implementations greatly reduce the time of training a large CNN;
(3) CNNs greatly improve the model generation capacity by introducing effective regularisation strategies, 
such as dropout~\cite{dropout}.

Despite the promising performance achieved by CNNs, 
it remains unclear how to design a `good' CNN architecture to adapt to a specific classification task 
due to the lack of theoretical guidance. 
However, some insights into CNN design can be gained by experimental comparisons of different CNN architectures.
The work~\cite{chatfield2014return}  made such comparisons and comprehensive analysis for the task of object recognition. 
However, face recognition is very  different from object recognition. 
Specifically, faces are aligned via 2D similarity transformation or 3D pose correction to a fixed reference position in images before feature extraction 
while object recognition usually does not conduct such alignment, and therefore objects appear in arbitrary positions. 
As a result,  the CNN architectures used for face recognition~\cite{DEEPID,DEEPID2,DEEPID2P,FACEPP2015,deepface} are rather different from those for object recognition~\cite{AlexNet,VERYDEEPOX,GoogleNet,PRELU}.
For the task of face recognition, it is important to  make a systematic evaluation of 
the effect of different CNN design and implementation choices.
In addition, those published CNNs~\cite{DEEPID,FACEPP2015,deepface,WEBFACE}  are trained in different face databases, most of which are not publicly available. 
The difference of training sets might result in unfair comparisons of CNN architectures. 
To avoid this unfairness, the comparison of different CNNs should be conducted on a common ground.

To  clarify the contributions of different components of  CNN-based face recognition systems, in this paper, a systematic evaluation is conducted.
To make our work reproducible, all the networks evaluated are trained on the publicly available LFW database. Specifically, our contributions are as follows: 

\begin{itemize}
\item Different CNN architectures including number of filters and layers are compared.  
      In addition, we evaluate the impact of multiple network fusion introduced by ~\cite{DEEPID}.
\item Various implementation choices, such as data augmentation, pixel value type (colour or grey) and similarity, are evaluated.
\item We quantitatively analyse how downstream metric learning methods such as joint Bayesian~\cite{JB} 
      can boost the effectiveness of the CNN-learned features.
\item Finally, source code for our CNN architectures and trained networks will be made publicly available (the training data is already public). This provides an extremely competitive baseline for face recognition to the community. To our knowledge, we are the first to publish fully reproducible CNNs for face recognition. 
\end{itemize}


\section{Related Work}
\label{RW}
CNN methods have drawn considerable attention in the field of face recognition in recent years.
In particular, CNNs have achieved impressive results on FRUE.
In this section, we briefly review these CNNs.
\begin{table*}[hpt] \small
\caption{Comparisons of 3 Published CNNs}

 \begin{threeparttable}
 \centering
\begin{tabular}{|c|c|c|c|c|c|c|}
\hline
           & Input      Image \tnote{1}                                                           & Architecture    \tnote{2}                                                                                                                                                                                      & No. of para. & \begin{tabular}[c]{@{}l@{}}Patch \\ Fusion\end{tabular} & \begin{tabular}[c]{@{}l@{}}Feature \\ Length\end{tabular}    & Training set                                                                    \\ \hline
DeepFace~\cite{deepface}   & 152$\times$152$\times$3                                                             & \begin{tabular}[c]{@{}l@{}}C1:32$\times$11$\times$11\tnote{3}, M2, C3:16$\times$9$\times$9, \\ L4: 16$\times$9$\times$9, L5:16$\times$7$\times$7, L6:16$\times$5$\times$5, \\F7, F8\end{tabular}                                                                                      & 120M+       & No                                                         & 4096   & \begin{tabular}[c]{@{}l@{}}120M+ images\\ 4K+ subjects\end{tabular} \\ \hline
DeepID  ~\cite{DEEPID}   & \begin{tabular}[c]{@{}l@{}}39$\times$31$\times$\{3,1\}\\ 31$\times$31$\times$\{3,1\}\end{tabular} & \begin{tabular}[c]{@{}l@{}}C1:20$\times$4$\times$4, M2, C3:40$\times$3$\times$3, M4,\\ C5:60$\times$3$\times$3, M6, C7:80$\times$2$\times$2,\\ F8, F9\end{tabular}                                                                                             &    101M+         & Yes                                                     & 19200  & \begin{tabular}[c]{@{}l@{}}202K+ images\\ 10K+ subjects\end{tabular}    \\ \hline
WebFace ~\cite{WEBFACE} & 100$\times$100$\times$1                                                             & \begin{tabular}[c]{@{}l@{}}C1:32$\times$3$\times$3, C2:64$\times$3$\times$3, M3, \\ C4:64$\times$3$\times$3, C5:128$\times$3$\times$3, M6, \\ C7:96$\times$3$\times$3, C8:192$\times$3$\times$3, M9,\\ C10:128$\times$3$\times$3, C11:256$\times$3$\times$3, M12, \\ C13:160$\times$3$\times$3, C14:320$\times$3$\times$3, A15, \\F16, F17\end{tabular} & 5M+         & No                                                         &   320      & \begin{tabular}[c]{@{}l@{}}986K+ images\\ 10K subjects\end{tabular}    \\ \hline
                                                                                                                                                                                                                                                                                                                                                                                                                                    
\end{tabular}
 \begin{tablenotes}
 \small{
  \item[1] The input image is represented as width$\times$height$\times$channels. 1 and 3 mean grey  or RGB images respectively.
  \item[2] The capital letters C, M, L, A, F represent convolutional, max pooling, locally connected, average pooling and fully connected layers respectively. These capital letters are followed by the indices of CNN layers.
  \item[3] The number of filters and filter size are denoted as `num $\times$  size $\times$ size' 
  }
 \end{tablenotes}
   \end{threeparttable}
\label{tab:structures}
\end{table*}

The researchers in Facebook AI group trained an 8-layer CNN named DeepFace~\cite{deepface}. 
The first three layers are conventional convolution-pooling-convolution layers.
The subsequent three layers are locally connected, followed by 2 fully connected layers. 
Pooling layers make learned features robust to local transformations 
but result in missing local texture details. 
Pooling layers are important for object recognition since the objects in images are not well aligned.
However, face images are well aligned before training a CNN. 
It is claimed in ~\cite{deepface} that one pooling layer is a good balance between local transformation robustness and preserving texture details.
DeepFace is trained on the largest face database to-date 
which contains four million facial images of 4,000 subjects.
Another contribution of ~\cite{deepface} is the 3D face alignment. 
Traditionally,  face images are aligned using 2D similarity transformation before they are fed into  CNNs.
However, this 2D alignment cannot handle out-of-plane rotations.
To overcome this limitation, ~\cite{deepface} proposes a 3D alignment method using an affine camera model.

In ~\cite{DEEPID}, a CNN-based face representation, referred to as Deep hidden IDentity feature (DeepID), is proposed. 
Unlike DeepFace whose features are learned  by one single big CNN, DeepID is learned by training a collection of small CNNs (network fusion). 
The input of one single CNN is the crops/patches of facial images and the features learned by all CNNs are concatenated to form a powerful feature.
Both RGB and grey crops  extracted around  facial points are used to train the DeepID.
The length of DeepID is 2 (RGB and Grey images) $\times$ 60 (crops) $\times$ 160 (feature length of one network) = 19,200.
One small network consists of 4 convolutional layers, 3 max pooling layers and  2 fully connected layers shown in Table ~\ref{tab:structures}.
DeepID uses  identification information only to supervise the CNN training. 
In comparison, DeepID2~\cite{DEEPID2}, an extension  of DeepID, uses both identification and verification information to train a CNN, 
aiming to maximise the inter-class difference but minimise the intra-class variations.
To further improve the performance of DeepID and DeepID2, DeepID2+~\cite{DEEPID2P} is proposed.
DeepID2+ adds the supervision information to all the convolutional layers rather than the topmost layers like DeepID and DeepID2. 
In addition, DeepID2+ improves the number of filters of each layer and uses a  much bigger training set than DeepID and DeepID2 .
In ~\cite{DEEPID2P}, it is also discovered that DeepID2+ has three interesting properties: being sparse, selective and robust.

The work~\cite{WEBFACE} proposes another face recognition pipeline, refereed to as WebFace, which also learns the face representation using a CNN.
WebFace collects a database which contains around 10,000 subjects and 500,000 images and makes this database publicly available. 
Motivated by very deep architectures of \cite{VERYDEEPOX,GoogleNet}, WebFace trains a much deeper CNN than those~\cite{DEEPID,DEEPID2,DEEPID2P,deepface} used for face recognition as shown in Table~\ref{tab:structures}.
Specifically, WebFace trains a 17-layer CNN which includes 10 convolutional layers, 5 pooling layers and 2 fully connected layers detailed in Table~\ref{tab:structures}.
Note that  the use of very small convolutional filters (3$\times$3), which avoids too much texture information decrease along a very deep architecture, is crucial to learn a powerful feature.
In addition, WebFace stacks two 3$\times$3 convolutional layers (without pooling in between) which is as effective as a 5$\times$5 convolutional layer
but with fewer parameters.

Table~\ref{tab:structures} compares three typical CNNs (DeepFace~\cite{deepface}, DeepID~\cite{DEEPID}, WebFace~\cite{WEBFACE}). 
It is clear that their architectures and implementation choices are rather different, which motivates our work.
In this study, we make  systematic evaluations to clarify the contributions of different components on a common ground.

\section{Methodology}
\label{sec:met}
LFW is the \emph{de facto} benchmark database for FRUE. 
Most exisiting CNNs~\cite{deepface, DEEPID, DEEPID2, DEEPID2P} train their networks on  private databases 
and test the trained models on LFW. 
In comparison, we train our CNNs only using LFW data  to  make our work easily reproducible. 
In this way, we cannot directly use  the reported CNN architectures~\cite{deepface, DEEPID, DEEPID2, DEEPID2P,WEBFACE} 
since our training data is much less extensive. 
We introduce three architectures adapting to our training set in subsection~\ref{method:arc}. 
To further improve the  discrimination of CNN-learned features, metric learning method is usually used.
One metric learning method, Joint Bayesian model~\cite{JB}, is detailed in subsection~\ref{method:MeL}.


\subsection{CNN Architectures}
\label{method:arc}
How to design a `good' CNN architecture remains an open problem. 
Generally, the architecture depends on the size of training data.
Less data should drive a smaller network (fewer layers and filters) to avoid overfitting.
In this study, the size of our training data is much smaller than that used by the state of the art methods~\cite{deepface, DEEPID, DEEPID2, DEEPID2P, WEBFACE};
therefore, smaller architectures are designed.

We propose three CNN architectures adapting to the size of  training data in LFW.
These  architectures are of three different sizes: small (CNN-S), medium (CNN-M), and large (CNN-L). 
CNN-S and CNN-M have 3 convolutional layers and two fully connected layers, while CNN-M has more filters than CNN-S.
Compared with CNN-S and CNN-M, CNN-L has 4 convolutional layers. 
The activation function  we used is REctification Linear Unit (RELU)~\cite{AlexNet}. 
In our experiments, dropout~\cite{dropout} does not improve the preformance of our CNNs,  therefore, it is not applied to our networks.
Following ~\cite{DEEPID, WEBFACE}, softmax function is used in the last layer for predicting 
a single class of K (the number of subjects in the context of face recognition) mutually exclusive classes.
During training, the learning rate is set to  0.001 for three networks, and the batch size is fixed to 100.
Table ~\ref{tab:ourCNN} details these three architectures.

\begin{table}[h]
  \caption{Our CNN Architectures}
\begin{tabular}{|c|c|c|} 
\hline
CNN-S                                                                                      & CNN-M                                                                                      & CNN-L                                                                                             \\ \hline
\multicolumn{3}{|c|}{conv1}                                                                                                                                                                                                                                                                 \\ \hline
\begin{tabular}[c]{@{}c@{}}12 $\times$ 5 $\times$ 5\\ st. 1, pad 0 \\ x2 pool\end{tabular} & \begin{tabular}[c]{@{}c@{}}16 $\times$ 5 $\times$ 5\\ st. 1, pad 0\\ x2 pool\end{tabular}  & \begin{tabular}[c]{@{}c@{}}16 $\times$ 3 $\times$ 3\\ st. 1, pad 1\\ -\end{tabular}               \\ \hline
\multicolumn{3}{|c|}{conv2}                                                                                                                                                                                                                                                                 \\ \hline
\begin{tabular}[c]{@{}c@{}}24 $\times$ 4 $\times$ 4\\ st. 1, pad 0 \\ x2 pool\end{tabular} & \begin{tabular}[c]{@{}c@{}}32 $\times$ 4 $\times$ 4\\ st. 1, pad 0 \\ x2 pool\end{tabular} & \begin{tabular}[c]{@{}c@{}}16 $\times$ 3 $\times$ 3\\ st. 1, pad 1\\ x2 pool\end{tabular}         \\ \hline
\multicolumn{3}{|c|}{conv3}                                                                                                                                                                                                                                                                 \\ \hline
\begin{tabular}[c]{@{}c@{}}32 $\times$ 3 $\times$ 3\\ st. 2, pad 0 \\ x2 pool\end{tabular} & \begin{tabular}[c]{@{}c@{}}48 $\times$ 3 $\times$ 3\\ st. 2, pad 0 \\ x2 pool\end{tabular} & \begin{tabular}[c]{@{}c@{}}32 $\times$ 3 $\times$ 3 \\ st. 1, pad 1\\ x3 pool, st. 2\end{tabular} \\ \hline
\multicolumn{3}{|c|}{conv4}                                                                                                                                                                                                                                                                 \\ \hline
-                                                                                          & -                                                                                          & \begin{tabular}[c]{@{}c@{}}48 $\times$ 3 $\times$ 3 \\ st. 1, pad 1\\ x2 pool\end{tabular}        \\ \hline
\multicolumn{3}{|c|}{fully connected}                                                                                                                                                                                                                                                       \\ \hline
160                                                                                        & 160                                                                                     & 160                                                                                            \\ \hline
4000, softmax                                                                              & 4000, softmax                                                                                    & 4000, softmax                                                                                           \\ \hline
\end{tabular}
\small{

Convolutional layer is detailed in 3 sub-rows: the 1st indicates  the number of filters and filter size as `num $\times$  size $\times$ size';
the 2nd specifies the convolutional stride (`st.') and padding (`pad'); and the 3rd specifies the max-pooling downsampling factor.
For fully connected layers, we specify their dimensionality: 160 for feature length and 4000 for the number of class/subjects. 
Note that every 9 splits (training set) of LFW have different number of subjects, but all around 4000.}
\label{tab:ourCNN}
\end{table}

\subsection{Metric Learning}
\label{method:MeL}
Metric Learning (MeL), which aims to find a new metric to make two classes more separable, is often used for  face verification.
MeL is independent of the feature extraction process and any feature (hand-crafted and learning-based) can be fed into a MeL method. 
Joint Bayesian (JB)~\cite{JB} model is a well-known MeL method 
and it is the most widely used MeL method which is applied to the features learned by CNNs~\cite{DEEPID,DEEPID2,WEBFACE}.

JB models the face verification task as a Bayesian decision problem. 
Let $H_I$ and $H_E$ represent intra-personal (matched) and extra-personal (unmatched) hypotheses, respectively. 
Based on the MAP (Maximum a Posteriori) rule, the decision is made by:
\begin{equation}
r(x_1,x_2)=\text{log}\frac{P(x_1,x_2\mid H_I)}{P(x_1,x_2\mid H_E)}
\label{rx1x2}
\end{equation}
where $x_1$ and $x_2$ are features of one face pair.
It is assumed that $P(x_1,x_2\mid H_I)$ and $P(x_1,x_2\mid H_E)$ have Gaussian distributions $N(0,S_I)$ and $N(0,S_E)$, respectively.

Before discussing the way of computing $S_I$ and $S_E$, we first explain the distribution of a face feature.
A face $x$ is modelled by the sum of two independent Gaussian variables (identity $\mu$ and intra-personal variations $\varepsilon$):
\begin{equation}
x=\mu +\varepsilon
\label{eq:aface}
\end{equation}
$\mu$ and $\varepsilon$ follow two Gaussian distributions $N(0,S_\mu)$ and $N(0,S_\varepsilon)$, respectively. 
$S_\mu$ and $S_\varepsilon$ are two unknown covariance matrices and they are regarded as face prior.
For the case of two faces,  the joint distribution of $\{x_1,x_2\}$ is also assumed as a Gaussian with zero mean. 
Based on Eq.~(\ref{eq:aface}), the covariance of two faces is:
\begin{equation}
\text{cov}(x_1,x_2)=\text{cov}(\mu_1,\mu_2)+\text{cov}(\varepsilon_1,\varepsilon_2)
\end{equation}
Then $S_I$ and $S_E$ can be derived as:
\begin{equation}
S_I=\begin{vmatrix}
  S_\mu+S_\varepsilon &  S_\mu \\ 
 S_\mu &   S_\mu+S_\varepsilon
\end{vmatrix}
\end{equation}
and 
\begin{equation}
S_E=\begin{vmatrix}
  S_\mu+S_\varepsilon &  0 \\ 
0 &   S_\mu+S_\varepsilon
\end{vmatrix}
\end{equation}

Clearly, $r(x_1,x_2)$ in Eq.~(\ref{rx1x2}) only depends on $ S_\mu$ and $S_\varepsilon$, which are learned from data using an EM algorithm~\cite{JB}.

\section{Evaluation}
\label{sec:eva}
LFW contains 5,749 subjects and 13,233 images and the training and test sets are defined in ~\cite{LFW}.
For evaluation, LFW is divided into 10 predefined splits for 10-fold cross validation. 
Each time nine of them are used for model training and the other one (600 image pairs) for testing. 
LFW defines three standard protocols (\emph{unsupervised, restricted} and \emph{unrestricted}) to evaluate face recognition performance. 
\emph{`Unrestricted'} protocol is applied here because the information of both subject identities and matched/unmatched labels is used in our system. 
The face recognition rate is evaluated by mean classification accuracy and standard error of the mean.

\begin{figure}[t]
\begin{center}
  \includegraphics[trim =30mm 65mm 50mm 20mm, clip, width=1  \linewidth]{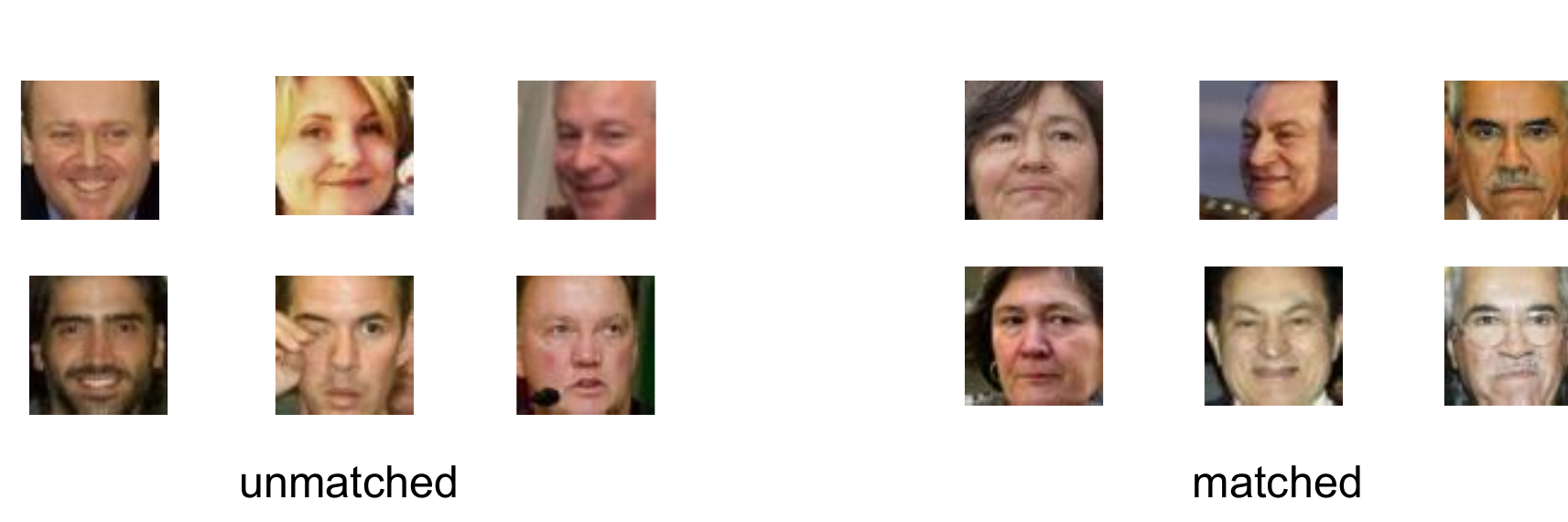}
\end{center}
   \caption{\footnotesize{Cropped sample  images in LFW}}
\label{fig:pairs}
\end{figure}

The images we used are aligned by  deep funneling~\cite{DeepFunneling}.
Each image is cropped to 58$\times$58 based on  the coordiates of two eye centers.
Some sample crops are visualised in Fig.~\ref{fig:pairs}.
It is commonly believed that data augmentation can boost the generalisation capacity of a neural network; therefore,  
each image is horizontally flipped.
The mean of the images is subtracted before network training.
The open source implementation MatConvNet~\cite{MatConvNet} is used to train our CNNs. 
In this section, different components of our CNN-based face recognition system are evaluated and analysed.

\paragraph {Architectures} 
Choosing a `good' architecture is crucial for CNN training. 
Overlarge or extremely small networks relative to the training data can lead to overfitting or underfitting, 
in the case of which the network does not converge at all during training. 
In this comparison, the RGB colour images are fed into CNNs and feature distance is measured by cosine distance.
The performances of the three architectures are compared in Table~\ref{tab:compArch}. 
CNN-M achieves the best face recognition performance, indicating that the  CNN-M generalises best among these three architectures using only LFW data. 
From this point, all the other evaluations are conducted using CNN-M. 
The face recognition rate 0.7882 of CNN-M is considered as the baseline, and all the remaining investigations will be compared with it.
\begin{table}[h]
\centering
\caption{Comparisons of Our Three Architectures}
\begin{tabular}{|c|c|}
\hline
Model & Accuracy      \\ \hline
CNN-S & 0.7828$\pm$0.0046 \\ \hline
CNN-M & 0.7882$\pm$0.0037 \\ \hline
CNN-L & 0.7807$\pm$0.0035 \\ \hline
\end{tabular}
\label{tab:compArch}
\end{table}

\vspace{-0.2cm}
\paragraph {Feature Distance}
The exisiting research offers little discussion about the distance measurement for CNN-learned features. 
In particular, it is interesting to know what is the best distance measure for face recognition. 
Table~\ref{tab:compDist}  compares the impact of six distance measures on face recognition accuracy. 
Cosine  and correlation achieve the best recognition rates, however, the standard deviation of cosine is smaller than that of   correlation.
Therefore, cosine distance is the best among these distances. 
\begin{table}[t]
\centering
\caption{Distance Comparison}
\begin{tabular}{|c|c|}
\hline
Distance    & Accuracy        \\ \hline
euclidean   & 0.6898$\pm$0.0092 \\ \hline
city block   & 0.6892$\pm$0.0088 \\ \hline
chebychev   & 0.6692$\pm$0.0088 \\ \hline
cosine      & 0.7882$\pm$0.0037 \\ \hline
correlation & 0.7882$\pm$0.0040 \\ \hline
spearman    & 0.7878$\pm$0.0031 \\ \hline
\end{tabular}
\label{tab:compDist}
\end{table}

\vspace{-0.2cm}
\paragraph {Grey vs Colour}
In  ~\cite{WEBFACE} and  ~\cite{deepface}, CNNs are trained using grey-level and RGB colour images, respectively. 
In comparison, both grey and colour images are used in ~\cite{DEEPID}.
We quantitatively compare the impact of these two  images types on face recognition.
Their comparative evaluation yields face recognition accuracies using grey and colour images of  0.7830$\pm$0.0077 and  0.7882$\pm$0.0118, respectively.
The performances using grey and colour images are very close to each other. 
Although  colour images contain more information, they do not deliver a significant improvement.

\vspace{-0.2cm}
\paragraph {Data Augmentation} Flip, mirroring images horizontally producing two samples from each, is a commonly used data augmentation technique for face recognition. 
Both original and mirrored images are used for training in all our evaluations. 
However,  little discussion in the existing work was made to analyse  the impact of image flipping during testing.  
Naturally, the test images can also be mirrored. 
A pair of test images can produce 2 new mirrored ones. These 4 images can generate 4 pairs instead of one original pair.
To combine these 4  images/pairs,  
two fusion strategies (feature and score fusion) are implemented  in this work. 
For feature fusion, the learned features of a test image and its mirrored one are concatenated to one feature, which is then used for score computing.
For score fusion, 4 scores generated from 4 pairs are averaged to one score. 
Table ~\ref{tab:compDA} compares the three scenarios: no flip during the test, feature and score fusions.
As is shown in Table ~\ref{tab:compDA}, mirroring images does improve the face recognition performance. 
In addition, feature fusion works slightly better than score fusion, however,  the improvements are not statistically significant.

\begin{table}[h]
\centering
\caption{Comparison of Data Augmentation during Test}
\begin{tabular}{|c|c|}
\hline
               & Accuracy            \\ \hline
 no flip on test set & 0.7882 $\pm$ 0.0037 \\ \hline
feature fusion & 0.7895 $\pm$ 0.0036 \\ \hline
score fusion   & 0.7893 $\pm$ 0.0035 \\ \hline
\end{tabular}
\label{tab:compDA}
\end{table}

\paragraph {Learned Feature Analysis} 
It is interesting to investigate the properties of CNN-learned face representations.
First, we discuss feature normalisation, which standardises the range of features and is generally performed during the data preprocessing step.
For example, to implement eigenface~\cite{eigenface}, the features (pixel values) are usually normalised via Eq.~(\ref{zscore}) before training a PCA space. 
\begin{equation}
\hat{\mathbf{x}}= \frac{\mathbf{x}-\mu_\mathbf{x}}{\sigma_\mathbf{x}}
\label{zscore}
\end{equation}
where $\mathbf{x} \in R$ and $\hat{\mathbf{x}} \in R$ are original and normalised feature vectors, respectively.
$\mu_\mathbf{x}$ and $\sigma_\mathbf{x}$ are the mean and standard deviation of $\mathbf{x}$. 
Motivated by this, our CNN features are normalised by Eq.~(\ref{zscore}) before computing cosine distance. 
The accuracies with and without  normalisation are 0.7927$\pm$0.0126  and  0.7882$\pm$0.0118, respectively. 
Thus  normalisation is  effective to improve recognition rate. 

\begin{figure}[t]
\begin{center}
  \includegraphics[trim =30mm 80mm 30mm 80mm, clip, width=0.9  \linewidth]{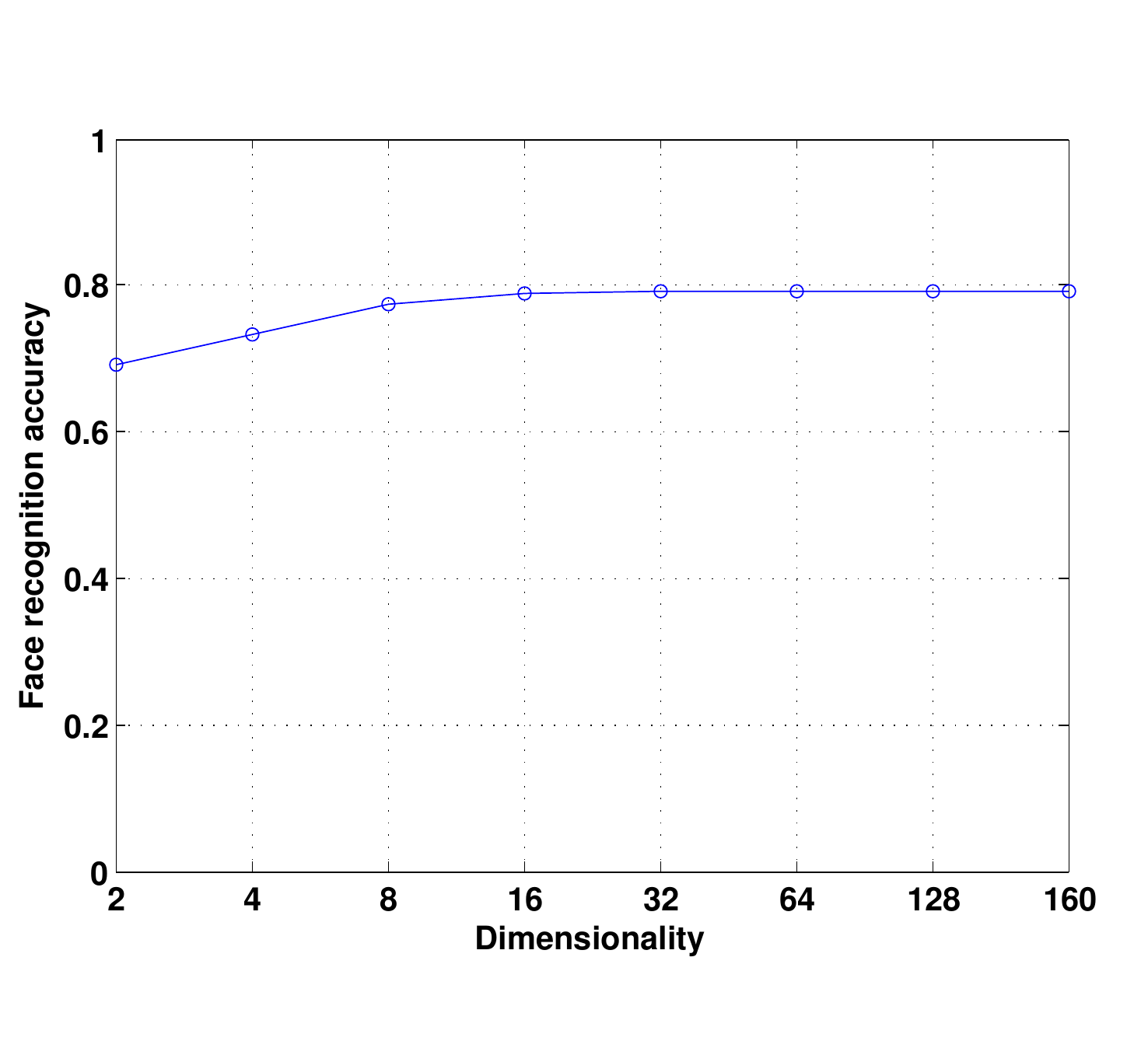}
\end{center}
   \caption{\footnotesize{The impact of feature dimensionality in PCA space on face recognition rate}}
\label{fig:pca}
\end{figure}

Second, we perform dimensionality reduction on the learned 160D features using PCA. 
As shown in Figure ~\ref{fig:pca},  only 16 dimensions of the   PCA feature space can achieve  comparable  face recognition rates 
to those of the original space. 
It is a very interesting property of CNN-learned features because low dimensionality can significantly reduce storage space and computation, which is crucial for large scale applications or mobile devices such as smartphone.

\paragraph {Network Fusion}
The work DeepID~\cite{DEEPID} and its variants~\cite{DEEPID2, DEEPID2P} apply the fusion of multiple networks. 
Specifically, the images of different facial regions and scales are separately fed 
into the networks that have the same architecture.  
The features learned from different networks are concatenated to a powerful face representation, 
which implicitly captures the spatial information of facial parts.
The size of these images can be different as shown in Table~\ref{tab:structures}. 
In ~\cite{DEEPID}, 120 networks are trained separately for this fusion. 
However, it is not very clear how greatly this fusion improves the face recognition performance. 
To clarify this issue, we implement the network fusion.

\begin{figure}[t]
\begin{center}
  \includegraphics[trim =30mm 100mm 20mm 80mm, clip, width=1  \linewidth]{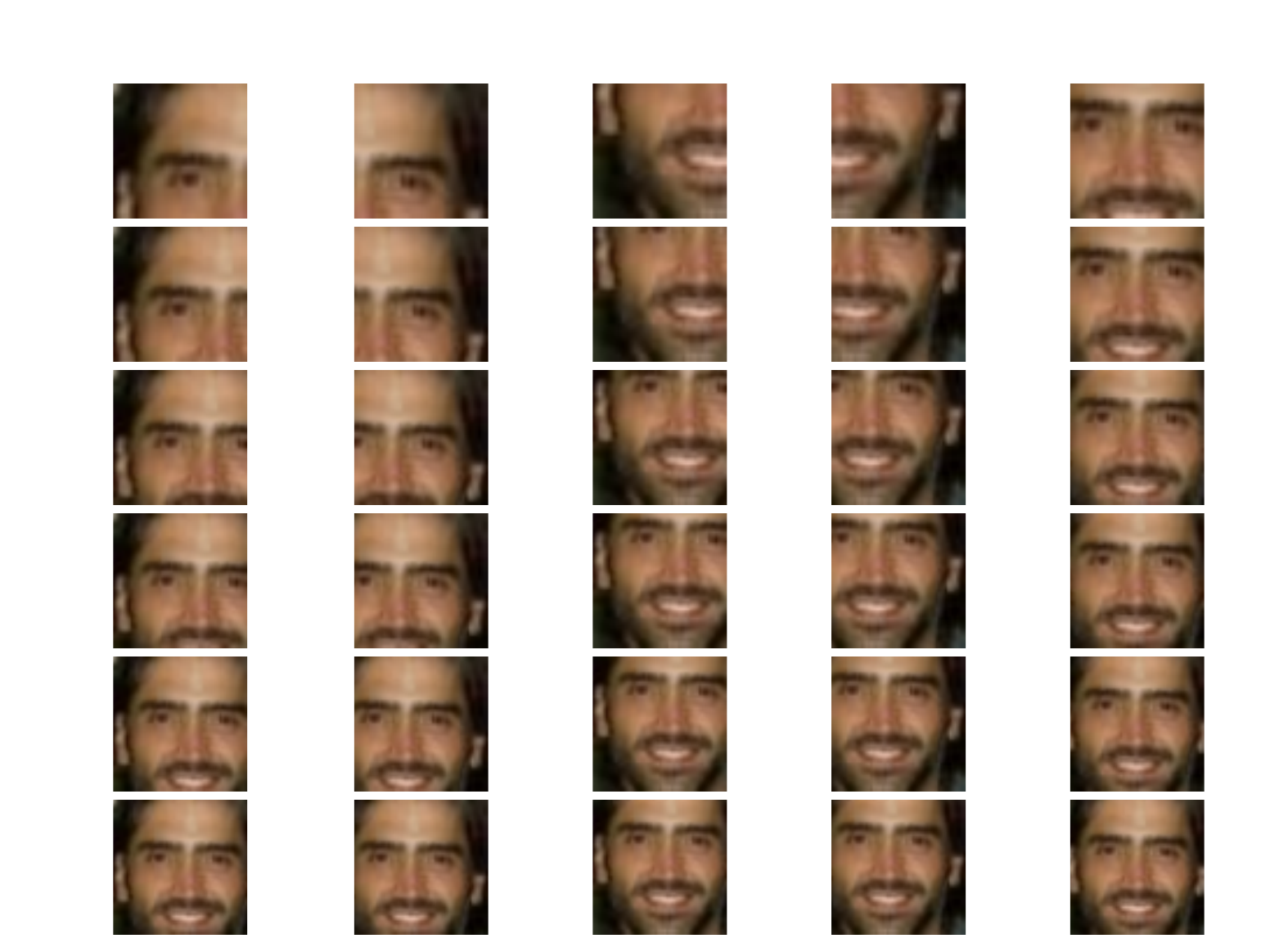}
\end{center}
   \caption{\footnotesize{Sample crops  in LFW. Rows correspond to  5 regions from 4 corners and center; Columns correspond to 6 scales.}}
\label{fig:30Region}
\end{figure}

We extract $d\times d$ crops from four corners and center and then upsample them to the original image size $58\times58$. 
The crops have 6 different scales: $ d= floor (58 \times \{ 0.3, 0.4, 0.5,0.6,0.7,0.8\})$, 
where  $floor$ is the operator to get the integer part.
Therefore we obtain 30 local patches with size of   $58\times58$ from one original image.
Figure~\ref{fig:30Region} shows these 30 crops.
To evaluate the performance of network fusion, we separately train  30 different networks using these crops. 
Then one face image can be represented by concatenating the features learned from different networks. 
Table ~\ref{tab:networkFus} compares the performance of single network and  network fusion. 
Note that we choose 16 best networks of 30 ones for the fusion.
It is clear that network fusion  works much better than a single network.
Specifically, the fusion of 16 best networks improves the face recognition accuracy of single network by 4.51\%. 
Clearly, the face representation of network fusion is actually the  fusion of features of different facial componets and scales. 
Similar ideas have widely been  used to improve the facial representation capacity of hand-crafted features 
such as multi-scale  local binary pattern~\cite{MLBP}, multi-scale local phase quantisation~\cite{MLPQ} 
and high-dimensional local features~\cite{HimF}.

\begin{table}[h]
\centering
\caption{Comparison of Network Fusion}
\begin{tabular}{|c|c|}
\hline
                 & Accuracy            \\ \hline
single network   & 0.7882 $\pm$ 0.0037 \\ \hline
network fusion & 0.8333 $\pm$ 0.0042 \\ \hline
\end{tabular}
\label{tab:networkFus}
\end{table}

\paragraph {Metric Learning}

\begin{figure}[t]
\begin{center}
  \includegraphics[trim =30mm 90mm 30mm 90mm, clip, width=1  \linewidth]{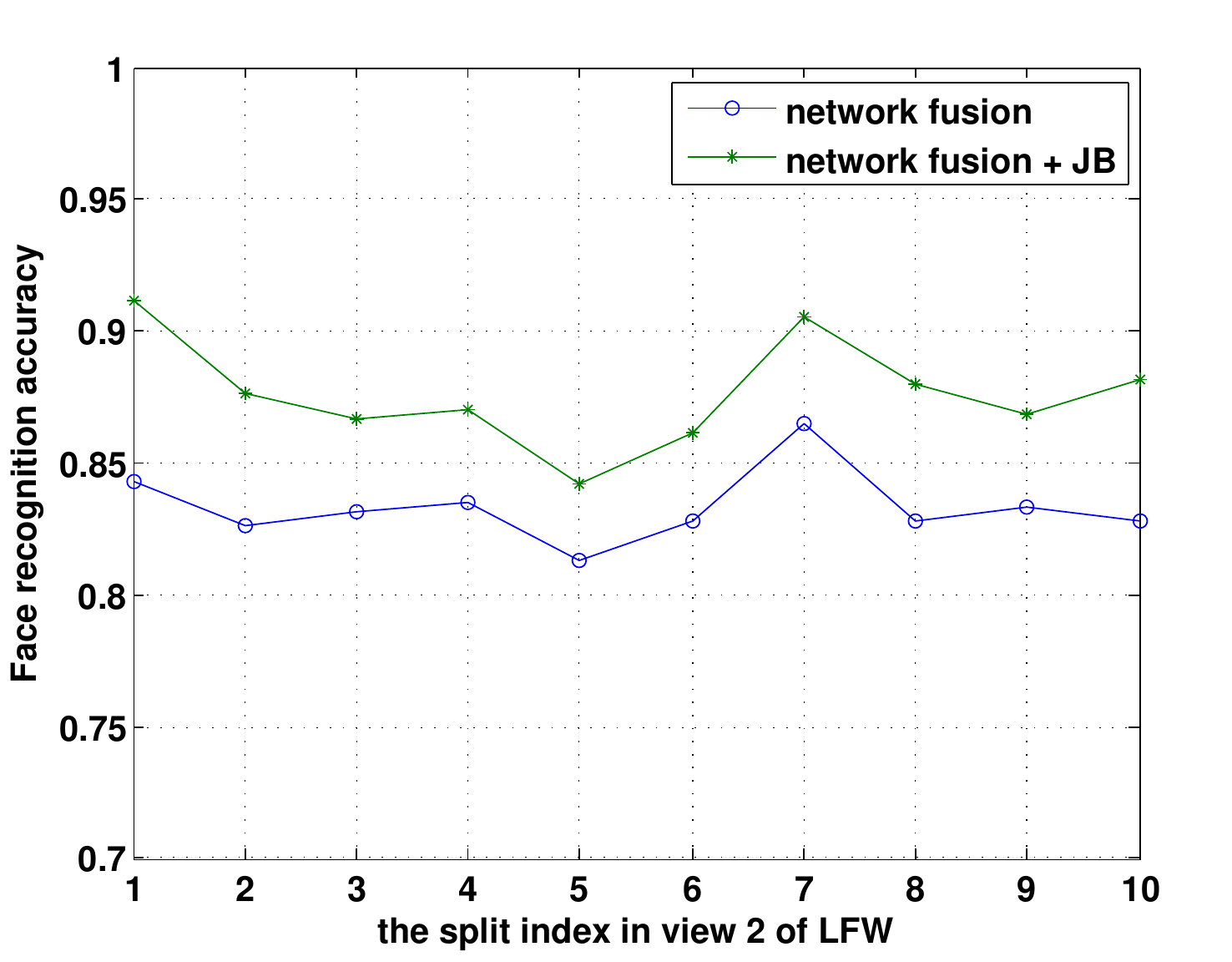}
\end{center}
   \caption{\footnotesize{Face recognition accuracies with and without JB }}
\label{fig:JB}
\end{figure}

For metric learning, the features of the fusion of  best 16 networks are used. 
The  feature dimensionality (2560=160$\times$16) is reduced to 320 via PCA before they are fed into  JB.
Figure~\ref{fig:JB} compares the face recognition accuracies with and without JB in each split of LFW database. 
JB  consistently and significantly improves the face recognition rates, showing the importance of metric learning.

Table~\ref{tab:onLFW} compares our method with non-commercial state-of-the-art methods. 
The performance of our method is slightly better than  ~\cite{LBPM,LDML,PLDA} but worse than~\cite{HimF, FVF, hybrid}.
However, the feature dimensionality of ~\cite{HimF, FVF} is much higher than ours.
In ~\cite{hybrid}, a large number of new pairs are generated in addition to those provided by LFW to train the model, while we do not generate new pairs.
\begin{table}[h]
\centering
\caption{Comparison with state-of-the-art methods on LFW under `unrestricted, label-free outside data'}
\begin{tabular}{|c|c|}
\hline
methods                     & accuracy                 \\ \hline
LBP multishot~\cite{LBPM}   & $0.8517 \pm 0.0061 $         \\ \hline
LDML-MkNN  ~\cite{LDML}     & $0.8750 \pm 0.0040 $         \\ \hline
LBP+PLDA~\cite{PLDA}        & $0.8733 \pm 0.0055  $        \\ \hline
high-dim LBP~\cite{HimF}    & $0.9318 \pm 0.0107$          \\ \hline
Fisher vector faces~\cite{FVF} & $0.9303 \pm 0.0105 $         \\ \hline
ConvNet+RBM  ~\cite{hybrid} & $0.9175 \pm 0.0048 $          \\ \hline
\textbf{Network fusion +JB} & $\textbf{0.8763} \pm \textbf{0.0064} $\\ \hline
\end{tabular}
\label{tab:onLFW}
\end{table}

\section{Conclusions} 
\label{sec:con}
Recently, convolutional neural networks have attracted a lot of attention in the field of face recognition. 
In this work, we present a rigorous empirical evaluation of CNN-based face recognition systems. 
Specifically, we quantitatively evaluate the impact of different architectures and implementation choices of CNNs on face recognition performances on common ground.
We have shown that network fusion can significantly improve the face recognition performance
because different networks capture the information from different regions and scales to form a powerful face representation. 
In addition, metric learning such as Joint Bayesian method can improve the face recognition greatly.

Since network fusion and metric learning are the two most important factors affecting  CNN  performance,
they will be the subject of future investigation. 

\paragraph{Acknowledgements} This work is supported by the European Union's Horizon 2020 research and innovation program under grant agreement No 640891, EPSRC/dstl project `Signal processing in a networked battlespace' under contract EP/K014307/1, EPSRC Programme Grant `S3A: Future Spatial Audio for Immersive Listener Experiences at Home' under contract EP/L000539, and the European Union project BEAT. We also gratefully acknowledge the support of NVIDIA Corporation for the donation of the GPUs used for this research.

{\small
\bibliographystyle{ieee}
\bibliography{egbib}
}

\end{document}